\documentclass[
]{ceurart}

\graphicspath{{figure/}}
\usepackage{subfigure}

\begin{document}

\copyrightyear{2020}
\copyrightclause{Copyright for this paper by its authors.\\
  Use permitted under Creative Commons License Attribution 4.0
  International (CC BY 4.0).}

\conference{Hate Speech and Offensive Content Identification in Indo-European Languages (HASOC) at Forum for Information Retrieval and Evaluation (FIRE),
  16th-20th December, 2020, Hyderabad, IN}

\title{HASOCOne@FIRE-HASOC2020: Using BERT and Multilingual BERT models for Hate Speech Detection}

\author[1]{Suman Dowlagar}[%
orcid=0000-0001-8336-195X,
]
\ead[1]{suman.dowlagar@research.iiit.ac.in}
\address[1]{International Institute of Information Technology - Hyderabad (IIIT-Hyderabad),
  Gachibowli, Hyderabad, Telangana, India, 500032}

\author[1]{Radhika Mamidi}
\ead[2]{radhika.mamidi@iiit.ac.in}

\begin{abstract}
  Hateful and Toxic content has become a significant concern in today's world due to an exponential rise in social media. The increase in hate speech and harmful content motivated researchers to dedicate substantial efforts to the challenging direction of hateful content identification.  In this task, we propose an approach to automatically classify hate speech and offensive content. We have used the datasets obtained from FIRE 2019 and 2020 shared tasks. We perform experiments by taking advantage of transfer learning models. We observed that the pre-trained BERT model and the multilingual-BERT model gave the best results. The code is made publically available at \url{https://github.com/suman101112/hasoc-fire-2020}
\end{abstract}


\maketitle

\section{Introduction}

Nowadays, people are frequently using social media platforms to communicate their opinions and share information. Although the communication among users can lead to constructive conversations, the people have been increasingly hit by hateful and offensive content due to these platforms' anonymity features. It has become a significant issue. The threat of abuse and harassment made many people stop expressing themselves. 

According to the Cambridge dictionary, Hate speech and offensive content is defined as,
\begin{itemize}
 \item To harass and cause lasting pain by attacking something uniquely dear to the target. 
 \item To use words that are considered insulting by most people.
\end{itemize}

The main obstacle with hate speech is, it is difficult to classify based on a single sentence because most of the hate speech has context attached to it, and it can morph into many different shapes depending on the context. Another obstacle is that humans cannot always agree on what can be classified as hate speech. Hence it is not very easy to create a universal machine learning algorithm that would detect it. Also, the datasets used to train models tend to "reflect the majority view of the people who collected or labeled the data".

To deal with the above scenarios and to encourage research on hate speech and offensive content, the NLP community organized several tasks and workshops such as Task 12: OffensEval 2: Multilingual Offensive content identification in Social Media text \footnote{https://sites.google.com/site/offensevalsharedtask/}, OSATC4 shared task on offensive content detection \footnote{http://edinburghnlp.inf.ed.ac.uk/workshops/OSACT4/}. Similarly, the FIRE 2020's HASOC shared task was devoted to the Hate Speech and Offensive Content Identification in Indo-European Languages. This task aims to classify the given annotated tweets. This paper presents the state-of-the-art BERT transfer learning models for automated detection of hate speech and offensive content. 

The paper is organized as follows. Section 2 provides related work on hate speech and offensive content detection.
Section 3 describes the methodology used for this task. Section 4 presents the experimental setup and the performance of the model. Section 5 concludes our work.

\section{Related Work}

Machine learning and natural language processing approaches have made a breakthrough in detecting hate speech on web platforms. Many scientific studies have been dedicated to using Machine Learning (ML) \cite{davidson2017automated,gaydhani2018detecting} and Deep Learning (DL)  \cite{gamback2017using,badjatiya2017deep} methods for automated hate speech and offensive content detection.  
The features used in traditional machine learning approaches are word-level and character-level n-grams, etc. Although supervised machine learning-based approaches have used different text mining-based features such as surface features, sentiment analysis, lexical resources, linguistic features, knowledge-based features, or user-based and platform-based metadata, they necessitate a well-defined feature extraction approach. Nowadays, the neural network models apply text representation and deep learning approaches such as Convolutional Neural Networks (CNNs) \cite{kim2014convolutional}, Bi-directional Long Short-Term Memory Networks (LSTMs) \cite{hochreiter1997long}, and BERT \cite{devlin2018bert} to improve the performance of hate speech and offensive content detection models.

\section{Methodology}

\begin{figure}
  \centering
  \includegraphics[scale=0.7]{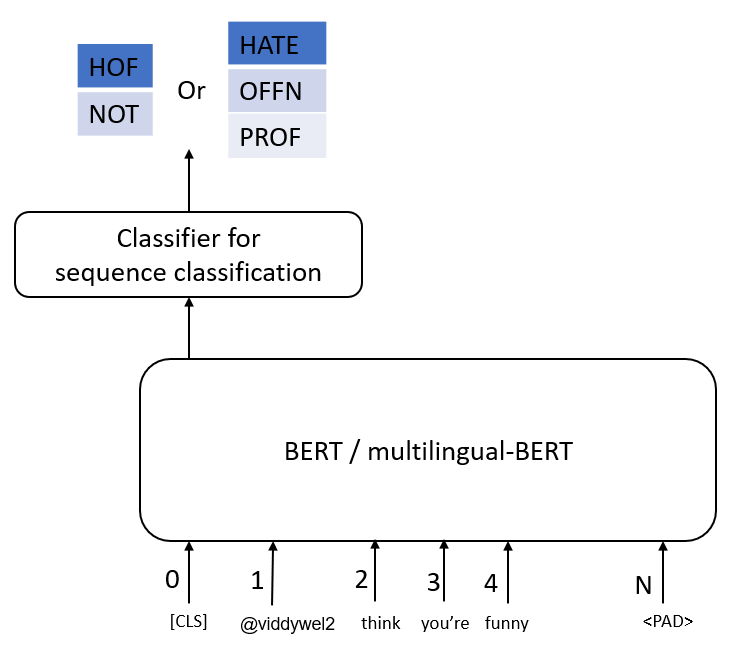}
  \caption{BERT model for sequence classification on Hate Speech Data.}
  \label{fig:model}
\end{figure}

Here, we use the pre-trained BERT transformer model for hate speech and offensive content detection. Figure \ref{fig:model} depicts the abstract view of BERT model that is used for hate speech detection and offensive language identification. Bidirectional Encoder Representations from Transformers (BERT) is a transformer Encoder stack trained on the large English corpus. It has 2 models, $BERT_{base}$ and $BERT_{large}$. These model sizes have a large number of transformer layers. The $BERT_{base}$ version has 12 transformer layers and the $BERT_{large}$ has 24. These also have larger feed-forward networks with 768 and 1024 hidden representations, and attention heads are 12 and 16 for the respective models. Like the vanilla transformer model \cite{vaswani2017attention}, BERT takes a sequence of words as input. Each layer applies self-attention, passes its results through a feed-forward network, and then hands it off to the next encoder. Embeddings from $BERT_{base}$ have 768 hidden units. The BERT configuration model takes a sequence of words/tokens at a maximum length of 512 and produces an encoded representation of dimensionality 768. 

The pre-trained BERT models have a better word representation as they are trained on a large Wikipedia and book corpus. As the pre-trained BERT model is trained on generic corpora, we need to fine-tune the model for the downstream tasks. During fine-tuning, the pre-trained BERT model parameters are updated when trained on the labeled hate speech and offensive content dataset. When fine-tuned on the downstream sentence classification task, a very few changes are applied to the $BERT_{base}$ configuration. In this architecture, only the [CLS] (classification) token output provided by BERT is used. The [CLS] output is the output of the 12th transformer encoder with a dimensionality of 768. It is given as input to a fully connected neural network, and the softmax activation function is applied to the neural network to classify the given sentence. Thus, BERT learns to predict whether a tweet can be classified as a hate speech or offensive content. Apart from $BERT_{base}$ model, we used the pre-trained multilingual $BERT_{base}$ model, as our data consisted of German and Hindi multilingual languages. The multilingual BERT and vanilla BERT models' architecture is the same, but the pre-trained multilingual BERT model is trained on multilingual Wikipedia language sources.

\section{Experiment}

Initially, we introduce datasets used, the task description, and then review the BERT model's performance on hate speech and offensive content detection. We also include our implementation details and error analysis in the subsequent sections.

\subsection{Dataset}

\begin{table*}
\centering
  \caption{Data Statistics}
  \label{tab:data}
\begin{tabular}{l|l|l}
\hline
\textbf{Language}             & \textbf{Train Sentences} & \textbf{Test Sentences} \\
\hline
English (HASOC 2019) & 5852               & 1153              \\
German (HASOC 2019)  & 3819               & 850               \\
Hindi (HASOC 2019)   & 4665               & 1318   
\\
\hline
English (HASOC 2020) & 3708               & 814               \\
German (HASOC 2020)  & 2373               & 526               \\
Hindi (HASOC 2020)   & 2963               & 663
\\
\hline
\end{tabular}
\end{table*}

We used the dataset provided by the organizers of HASOC FIRE-2020 \cite{hasoc2020overview} and FIRE-2019 \cite{mandl2019overview}. The HASOC dataset was subsequently sampled from Twitter and partially from Facebook for \textit{English}, \textit{German}, and \textit{Hindi} languages. The tweets were acquired using hashtags and keywords that contained offensive content. The statistics of FIRE 2020 and 2019 datasets are given in the Table \ref{tab:data}. 

\subsection{Task description}

The following tasks are in HASOC 2020.

\textbf{Sub-task A} focuses on coarse-grained Hate speech detection in all three languages. The task is to classify tweets into two classes: 
\begin{itemize}
 \item (NOT) Non Hate-Offensive - Post does not contain any Hate speech, profane, offensive content.
 \item (HOF) Hate and Offensive - Post contains Hate, offensive, and profane content.
\end{itemize}

\textbf{Sub-task B} represents a fine-grained classification. Hate-speech
and offensive posts from the sub-task A are further classified into three categories. The task is to classify the tweets into three classes:
\begin{itemize}
 \item (HATE) Hate speech - Post contains Hate speech content.
\item (OFFN) Offenive - Post contains offensive content such as insulting, degrading, dehumanizing and threatening.
\item (PRFN) Profane - Post contains profane words. This typically concerns the usage of swearwords and cursing.
\end{itemize}

\subsection{Implementation}

For the implementation, we used the transformers library provided by HuggingFace \cite{wolf2019huggingface}. The HuggingFace transformers package is a python library providing pre-trained and configurable transformer models useful for a variety of NLP tasks. It contains the pre-trained BERT and multilingual BERT, and other models suitable for downstream tasks. As the implementation environment, we use the PyTorch library that supports GPU processing. The BERT models were run on NVIDIA RTX 2070 graphics card with an 8 GB graphics card. We trained our classifier with a batch size of 64 for 5 to 10 epochs based on our experiments. The dropout is set to 0.1, and the Adam optimizer is used with a learning rate of 2e-5. We used the hugging face transformers pre-trained BERT tokenizer for tokenization. We used the BertForSequenceClassification module provided by the HuggingFace library during finetuning and sequence classification. 

\subsection{Baseline models}

Here, we compared the BERT model with other machine learning algorithms.

\subsubsection{SVM with TF\_IDF text representation}

We chose Support Vector Machines (SVM) for hate speech and offensive content detection. The tokenizer used is SentencePiece \cite{kudo2018sentencepiece}. SentencePiece is a commonly used technique to segment words into a subword-level. In both cases, the vocabulary is initialized with all the individual characters in the language, and then the most frequent or likely combinations of the symbols are iteratively added to the vocabulary.

\subsubsection{ELMO embeddings with SVM model}

ELMO(Embeddings from Language Models)  \cite{peters2018deep} deals with contextual embeddings. Contextual word-embeddings are born to capture the word meaning in its context. Instead of using a fixed embedding for each word, ELMO looks at the word's context, i.e., the word's entire sentence, before assigning embedding to the word. It uses a bi-LSTM trained on a specific task to be able to create those embeddings. We used the ELMO model present on tensorflow hub (\url{https://tfhub.dev/google/elmo/2}) to obtain the ELMO embeddings on the hate speech data for all the languages. After obtaining the embeddings, we take the mean of embeddings and apply an SVM classifier to classify the given sentence into hate speech or offensive content. We used the SentencePiece tokenizer.

\section{Results}

\begin{table}[]
\caption{macro F1 and Accuracy on English Subtasks A and B}
\label{tab:en_result}
\begin{tabular}{l|cc|cc}
\hline
                    & \multicolumn{2}{l|}{\textbf{Hate speech Detection}} & \multicolumn{2}{l}{\textbf{Offensive Content Identification}} \\
\hline
\textbf{Model}      & \textbf{macro F1}        & \textbf{Accuracy}       & \textbf{macro F1}             & \textbf{Accuracy}             \\
\hline
\textbf{SVM}        &   81.56\%                       & 81.57\%                        & 47.49\%                               & 76.78\%                             \\
\textbf{ELMO + SVM} &   82.43\%                       & 83.78\%                         & 49.62\%                              & 79.54\%                              \\
\textbf{BERT}       & 88.33\%                         & 88.33\%                         & 54.44\%                             & 81.57\%   \\
\hline
\end{tabular}
\end{table}

\begin{table}[]
\caption{macro F1 and Accuracy on German Subtasks A and B}
\label{tab:ge_result}
\begin{tabular}{l|cc|cc}
\hline
                    & \multicolumn{2}{l|}{\textbf{Hate speech Detection}} & \multicolumn{2}{l}{\textbf{Offensive Content Identification}} \\
\hline
\textbf{Model}      & \textbf{macro F1}        & \textbf{Accuracy}       & \textbf{macro F1}             & \textbf{Accuracy}             \\
\hline
\textbf{SVM}        &  73.29\%                        & 79.27\%                         &  45.54\%                             & 77.94\%                              \\
\textbf{ELMO + SVM} &  71.73\%                        & 80.42\%                        &  45.94\%                             & 78.21\%                              \\
\textbf{multilingual-BERT}       & 77.91\%                     &  82.51\%                       &  47.78\%                             & 80.42\%   \\
\hline
\end{tabular}
\end{table}

\begin{table}[]
\caption{macro F1 and Accuracy on Hindi Subtasks A and B}
\label{tab:hi_result}
\begin{tabular}{l|cc|cc}
\hline
                    & \multicolumn{2}{l|}{\textbf{Hate speech Detection}} & \multicolumn{2}{l}{\textbf{Offensive Content Identification}} \\
\hline
\textbf{Model}      & \textbf{macro F1}        & \textbf{Accuracy}       & \textbf{macro F1}             & \textbf{Accuracy}             \\
\hline
\textbf{SVM}        &   59.73\%                       & 70.13\%                         & 36.78\%                              & 72.39\%                             \\
\textbf{ELMO + SVM} &   60.91\%                       & 71.47\%                        &  39.89\%                             & 72.76\%                              \\
\textbf{multilingual-BERT}       & 63.54\%                         & 74.96\%                         &  49.71\%                             & 73.15\%   \\
\hline
\end{tabular}
\end{table}

The results are tabulated in Tables \ref{tab:en_result}, \ref{tab:ge_result} and \ref{tab:hi_result}. We evaluated the performance of the method using macro F1 and accuracy. The BERT model performed well when compared to the other SVM with TF-IDF and ELMO text representations. Given all the languages and both the subtasks A and B, we have observed an increase of 1-2\% in classification metrics for ELMO embeddings + SVM classifier compared to the baseline SVM classifier. However, BERT showed an increase of 5-7\% in classification metrics compared to ELMO and SVM models. It shows the pre-trained BERT model's capability, which learnt better text representations from the generic data. The state of the art transformer architecture used in the BERT model helped the model learn better parameter weights in hate speech and offensive content detection. 

\begin{figure}[pos=!t]
\centering
\subfigure[]{\label{cm_ea}\includegraphics[width=63mm]{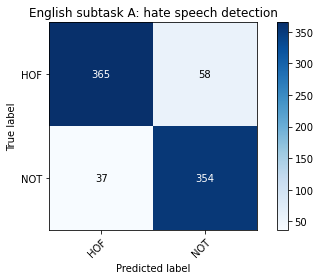}}
\subfigure[]{\label{cm_eb}\includegraphics[width=63mm]{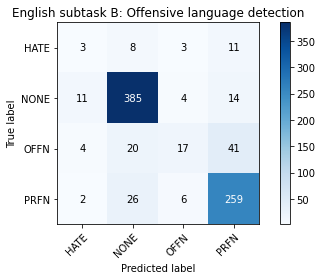}}
\subfigure[]{\label{cm_ga}\includegraphics[width=63mm]{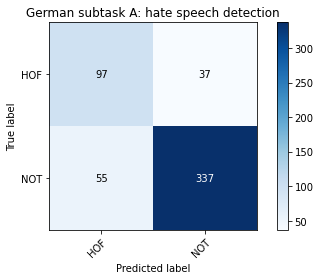}}
\subfigure[]{\label{cm_gb}\includegraphics[width=63mm]{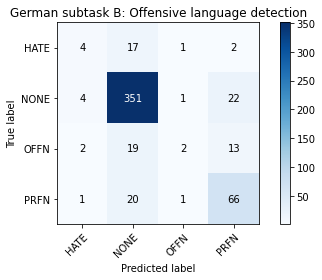}}
\subfigure[]{\label{cm_ha}\includegraphics[width=63mm]{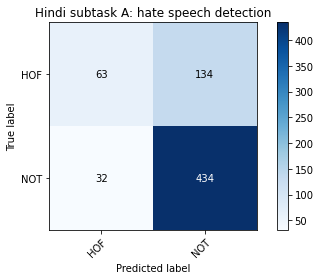}}
\subfigure[]{\label{cm_hb}\includegraphics[width=63mm]{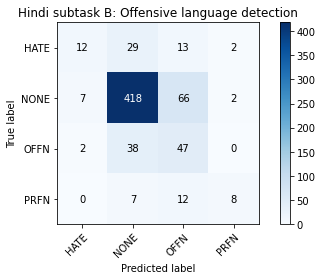}}
\caption{Confusion matrix on the given test data for the English, German and Hindi languages given subtask A: Hate Speech Detection and subtask B: Offensive Content Identification}
\label{fig:cm}
\end{figure}

\section{Error Analysis}

The confusion matrix of BERT model for subtasks A and B for the english, german and hindi datasets is given in the Figure \ref{fig:cm}. 
For the binary classification, the best-performed model was for English subtask A. The binary classification for the Hindi model is not helpful. The model misclassified most of the hate-speech labels. It can be seen in subfigure \ref{cm_ha}.
For offensive content evaluation, the model performed better on English subtask B. It correctly classified "NONE (not offensive)" and "PROF (profane)" but was unable to classify "HATE (hate speech)" and "OFFN (offensive)" and misunderstood most of them as "PROF". The multilingual-BERT model misclassified most of the hate speech and offensive content labels for the German and Hindi languages as "NONE" and didn't perform well on those datasets.

\section{Conclusion and Future work}
We used pre-trained bi-directional encoder representations using transformers (BERT) and multilingual-BERT for hate speech and offensive content detection for English, German, and Hindi languages. We compared the BERT with other machine learning and neural network classification methods. Our analysis showed that using the pre-trained BERT and multilingual BERT models and finetuning it for downstream hate-speech text classification tasks showed an increase in macro F1 score and accuracy metrics compared to traditional word-based machine learning approaches.

The given data has both hate speech and offensive content labeled for a given same sentence. It implies that both tasks are related. In such a scenario, we can use joint learning models to help obtain a strong relationship between the two tasks. Which, in turn, helps a deep joint classification model to understand the given datasets better.

\bibliography{sample-ceur}


\end{document}